\documentclass[runningheads]{llncs}
\usepackage{marvosym}
\usepackage{graphicx}
\usepackage{comment}
\usepackage{amsmath,amssymb} 
\usepackage{color}
\usepackage{multicol}
\usepackage{booktabs}       
\usepackage{multirow}
\usepackage{wrapfig}
\usepackage{subfigure}

\begin{document}
\pagestyle{headings}
\mainmatter
\def\ECCVSubNumber{2916}  

\title{Accurate RGB-D Salient Object Detection via Collaborative Learning} 

\titlerunning{Accurate RGB-D Salient Object Detection via Collaborative Learning}
%
\author{Wei Ji\thanks{means equal contribution.}\inst{1}\and
Jingjing Li$^\star$\inst{1}\and
Miao Zhang\textsuperscript{\Letter}\inst{1} \and
Yongri Piao\inst{1}\and
Huchuan Lu\inst{1,2}}
\authorrunning{W. Ji, et al.}
%
\institute{Dalian University of Technology, Dalian, China \and
Pengcheng Lab, Shenzhen,China\\
\email{weiji.dlut@gmail.com, jingjing.dlut@outlook.com},\\
\email{\{miaozhang, yrpiao, lhchuan\}@dlut.edu.cn}\\
\url{https://github.com/OIPLab-DUT/CoNet}}
\maketitle

\begin{abstract}
Benefiting from the spatial cues embedded in depth images, recent progress on RGB-D saliency detection shows impressive ability on some challenge scenarios.
However, there are still two limitations.
One hand is that the pooling and upsampling operations in FCNs might cause blur object boundaries.
On the other hand, using an additional depth-network to extract depth features might lead to high computation and storage cost.
The reliance on depth inputs during testing also limits the practical applications of current RGB-D models. 
In this paper, we propose a novel collaborative learning framework where edge, depth and saliency are leveraged in a more efficient way, which solves those problems tactfully.
The explicitly extracted edge information goes together with saliency to give more emphasis to the salient regions and object boundaries.
Depth and saliency learning is innovatively integrated into the high-level feature learning process in a mutual-benefit manner.
This strategy enables the network to be free of using extra depth networks and depth inputs to make inference.
To this end, it makes our model more lightweight, faster and more versatile.
Experiment results on seven benchmark datasets show its superior performance.
\end{abstract}

\section{Introduction}
The goal of salient object detection (SOD) is to locate and segment the most attractive and noticeable regions in an image.
As a fundamental and pre-processing task, salient object detection plays an important role in various computer vision tasks, e.g., visual tracking~\cite{Track1,Track2}, video SOD~\cite{videoSOD1,videoSOD2}, object detection~\cite{Recon1,Recon2}, semantic segmentation~\cite{Segment}, and human-robot interaction~\cite{Robot}.
\begin{figure}[!htbp]
\centering 
\includegraphics [width=1\linewidth] {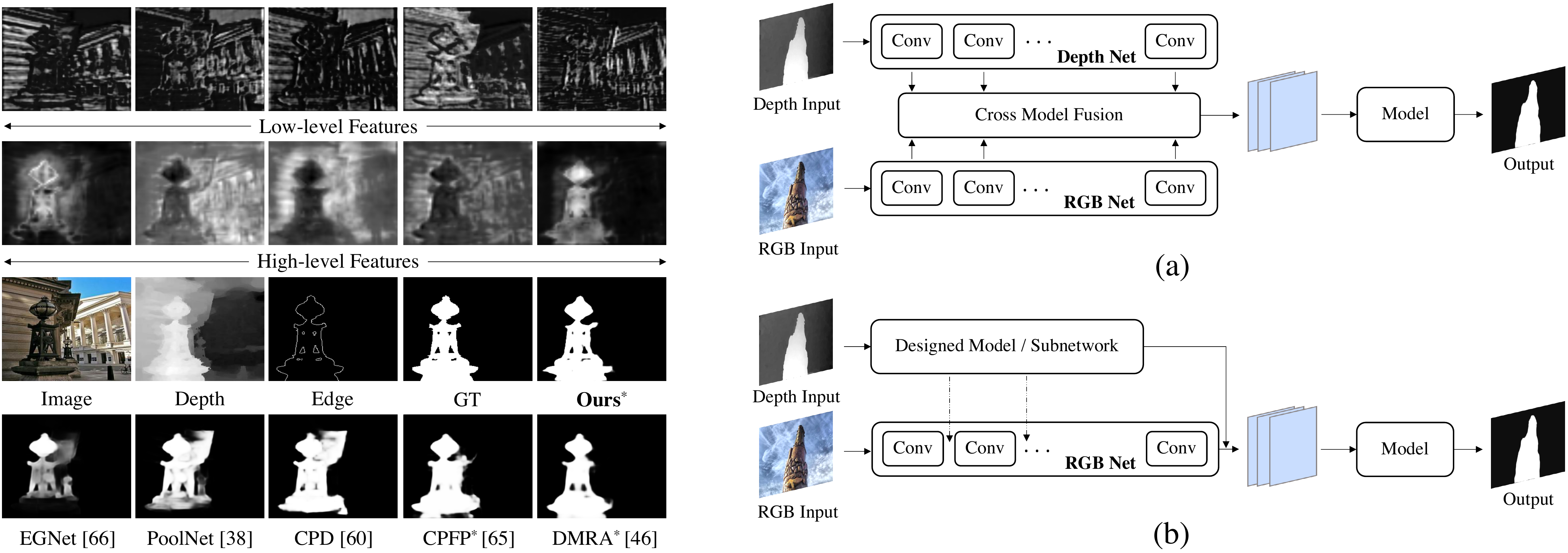}
\vspace{-0.5cm}
\caption{\textbf{(Left)} First two rows: feature maps in different layers of CNNs. Last two rows: RGB image, depth image, edge map, saliency ground truth (GT) and saliency results of several state-of-the-art methods. * means RGB-D methods. \textbf{(Right)} Two kinds of previous RGB-D SOD network structures. (a) Processing RGB input and depth input separately and then combining the complementary RGB and depth features through cross-modal fusion (e.g.~\cite{3DMPCI,3DCTMF,3DPCA,3DTANet,3DDMRA}). (b) Using tailor-made depth subnetworks to compensate for RGB representations (e.g.~\cite{3DPDNet,3DCPFP}).}
\vspace{-0.4cm}
\label{fig:introduction}
\end{figure}

Recent researches on RGB-D salient object detection have gradually broken the performance bottleneck of traditional methods and RGB-based methods, especially when dealing with complex scenarios like similar foreground and background.
However, there are some limitations with the introduction of FCNs~\cite{FCN,VGG} and depth images.
\emph{Firstly}, the emergence of FCNs enables automatic extraction of multi-level and multi-scale features.
The high-level features with rich semantic information can better locate salient objects but the pooling and upsampling operations in FCNs might result in coarse and blur object boundaries (see Fig.~\ref{fig:introduction} (left)).
The low-level features contain rich local details but suffer from excessive background noises and might cause information chaos.
\emph{Secondly}, the spatial layout information from depth images can better express 3D scenes and help locate salient objects.
However, previous RGB-D methods either adopted two-stream architectures that process RGB and depth images separately with various cross-modal fusion strategies (see Fig.~\ref{fig:introduction}a)~\cite{3DMPCI,3DCTMF,3DPCA,3DTANet,3DDMRA},
or utilized subnetworks tailored for depth image to compensate for RGB representations (see Fig.~\ref{fig:introduction}b)~\cite{3DPDNet,3DCPFP}.
In those methods, the additional depth-networks might lead to high computation and storage cost, and cannot work without depth input, seriously limiting their practical applications.

In this paper, we propose a novel collaborative learning framework (CoNet) to confront the aforementioned limitations.
In collaborative learning, multiple group members work together to achieve learning goals through exploratory learning and timely interaction.
In our framework, three mutually beneficial collaborators are well-designed from different perspectives of the SOD task, namely edge detection, coarse salient object detection, and depth estimation.
\emph{On the one hand}, a edge collaborator is proposed to explicitly extracts edge information from the overabundant low-level features and then goes together with saliency knowledge to jointly assign greater emphasis to salient regions and object boundaries.
\emph{On the other hand}, considering the strong consistencies among global semantics and geometrical properties of image regions~\cite{Towards},
we innovatively integrate depth and saliency learning into the high-level feature learning process in a mutual-benefit manner.
Instead of directly taking depth image as input,
this learning strategy enables the network to be free of using an extra depth network to make inference from an extra input.
Compared with previous RGB-D models which utilize additional subnetworks to extract depth features and rely on depth images as input, our network is more lightweight, faster and more versatile.
To our best knowledge, this is the first attempt to use depth images in such a way in RGB-D SOD research.
\emph{Finally}, a unified tutor named knowledge collector is designed to accomplish knowledge transfer from individual collaborators to the group, so as to more comprehensively utilize the learned edge, saliency and depth knowledges to make accurate saliency prediction.
Benefiting from this learning strategy, our framework produces accurate saliency results with sharp boundary preserved and simultaneously avoids the reliance on depth images during testing.

In summary, our main contributions are as follows:
\begin{itemize}
\setlength{\itemsep}{1pt}
\setlength{\parsep}{0pt}
\setlength{\parskip}{0pt}
\item We propose a novel collaborative learning framework (CoNet) where edge, depth, and saliency are leveraged in a different but more efficient way for RGB-D salient object detection.
The edge exploitation makes the boundaries of saliency maps more accurate.
\item This learning strategy enables our RGB-D network to be free of using an additional depth network and depth input during testing, and thus being more lightweight and versatile.
\item Experiment results on seven datasets show the superiority of our method over other state-of-the-art approaches.
Moreover, it supports the faster frame rate as it runs at 34 FPS, meeting the needs of real-time prediction (enhances FPS by 55\% compared with current best performing method DMRA~\cite{3DDMRA}). 
\end{itemize}

\section{Related Work}
Early works~\cite{2DT1,2DT2,2DT6,2DData1,2DData2} for saliency detection mainly rely on hand-crafted features.
\cite{2DF,Benchmark,2DDeep} are some comprehensive surveys.
Recently, traditional methods have been gradually surpassed by deep learning ones.
Among those researches, 2D methods~\cite{2DLi,2DData3,2DWang,2DZhao,2DLee,2DT8,2DwangDVAP,2DwangFP,2DSOC,2DNLDF,2DCPD} based on RGB images have achieved remarkable performance and lone been the mainstream of saliency detection.
However, 2D saliency detection appears to make a downgrade when handling complex scenarios due to the lack of spatial information in single RGB image.
The introduction of depth images in RGB-D saliency researches~\cite{3DGP,3DTPPF,3DDF,3DCTMF,3DMPCI,3DPCA,3DTANet,3DPDNet,3DDMRA,3DCPFP} has made great promotions for those complex cases thanks to the embedded rich spatial information of depth images.

The first CNNs-based method~\cite{3DDF} for RGB-D SOD uses hand-crafted features extracted from RGB and depth images for training.
Then, Chen \emph{et~al.} propose to use two-stream models~\cite{3DCTMF,3DMPCI} to process RGB and depth image separately and then combine cross-modal features to jointly predict saliency.
They subsequently design a progressive fusion network~\cite{3DPCA} to better fuse cross-modal multi-level features and propose a three-stream network~\cite{3DTANet} which adopts the attention mechanism to adaptively select complement from RGB and depth features.
Afterwards,
Piao \emph{et~al.}~\cite{3DDMRA} utilize residual structure and depth-scale feature fusion module to fuse paired RGB and depth features. 
The network structures in~\cite{3DCTMF,3DMPCI,3DPCA,3DTANet,3DDMRA} can be represented as two-stream architectures shown in Fig.~\ref{fig:introduction}a.
Another kind of structure is the use of subnetworks tailored for depth images to extract depth features and make compensation for RGB representations~\cite{3DPDNet,3DCPFP} (Fig.~\ref{fig:introduction}b).
Zhu \emph{et~al.}~\cite{3DPDNet} utilize an auxiliary network to extract depth-induced features and then use them to enhance a pre-trained RGB prior model.
In~\cite{3DCPFP}, Zhao \emph{et~al.} first enhance the depth map by contrast prior and then think of it as an attention map and integrate it with RGB features.

Those methods have some limitations.
Using additional depth networks to extract depth features leads to high computation and storage cost.
The reliance on depth images as input during testing also severely limits the practical applications of current RGB-D models.
Moreover, we found that the boundaries of the produced saliency maps in those methods are a bit coarse and blur.
This is mainly because the pooling and upsampling operations in FCNs might lead to the loss of local details and current RGB-D methods have not taken steps to emphasize the boundaries of salient objects.

Some RGB-based SOD methods attempt to enhance the boundary accuracy through adding edge constraints or designing boundary-aware losses.
An edge guidance network~\cite{2DEGNet} couples saliency and edge features to better preserve accurate object boundary.
Liu \emph{et~al.}~\cite{2DPoolNet} train their pooling-based network with edge detection task and successfully enhance the details of salient regions.
A predict-refine architecture~\cite{2DBASNet} equipped with a hybrid loss segments salient regions and refines the structure with clear boundaries.
An attentive feedback module~\cite{2DAFNet} employs a boundary-enhanced loss for learning exquisite boundaries.

In this paper, we propose a novel collaborative learning framework where edge, depth and saliency are leveraged in a different but more efficient way.
Different from previous RGB methods using edge supervision~\cite{2DEGNet,2DPoolNet} or boundary-aware losses~\cite{2DBASNet,2DAFNet}, we further combine the learned edge knowledge with saliency knowledge to give extra emphasis to both salient regions and boundaries.
For the use of depth, we innovatively integrate it into the high-level feature learning process in a mutual-benefit manner, instead of directly taking depth images as input.
Free of using the depth subnetworks and depth input during testing makes our network more lightweight and versatile.
In Section~\ref{section:3}, we will elaborate on our collaborative learning framework.

\section{Collaborative Learning Framework}
\label{section:3}

\subsection{The Overall Architecture}

\begin{figure*}[t]
\centering 
\includegraphics [width=1\linewidth] {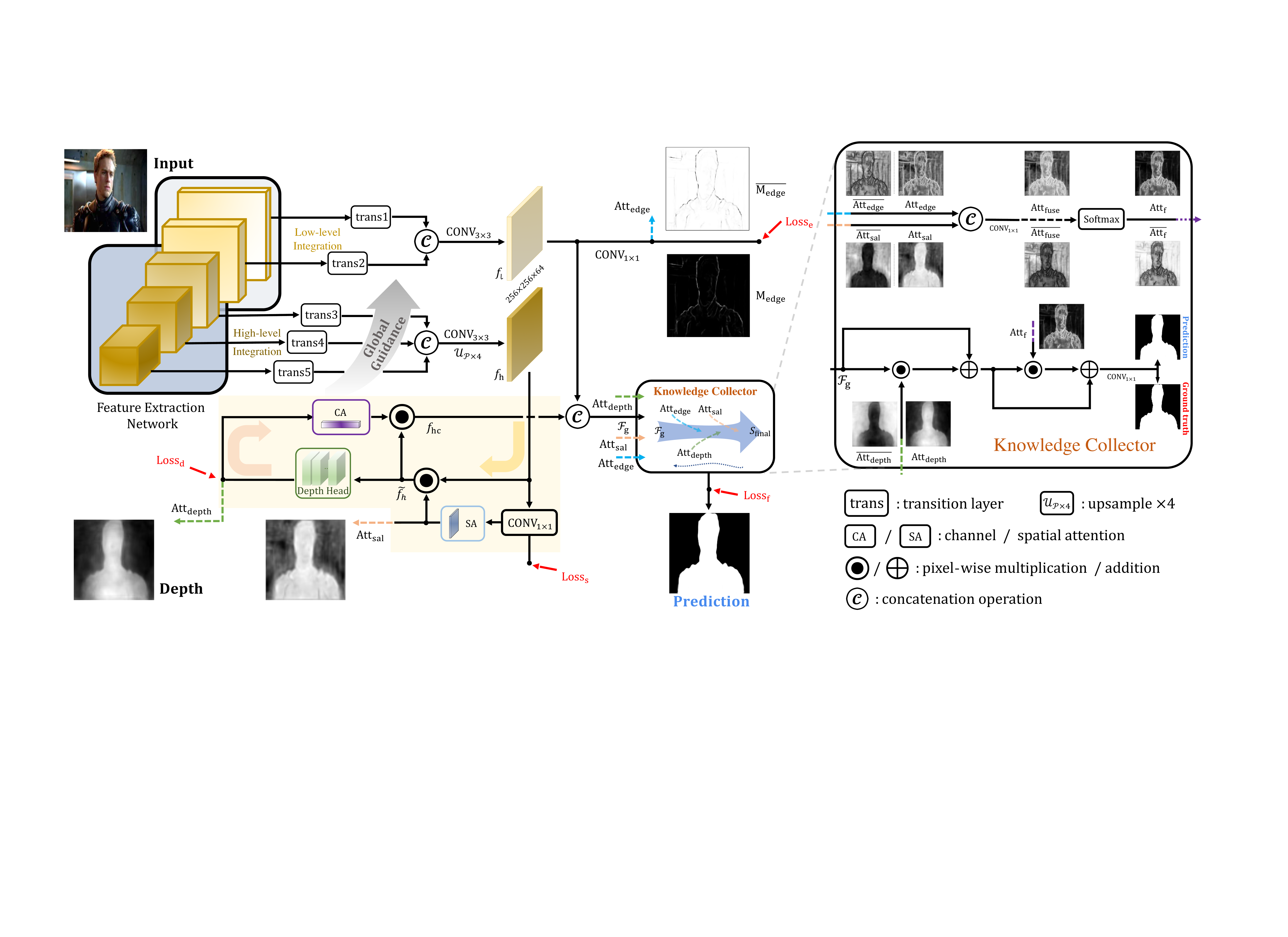}
\vspace{-0.55cm} 
\caption{The overall architecture of our collaborative learning framework. Details of the Global Guidance Module can be found in Fig.~\ref{fig:GGM}. Here, $\overline{Att_*} = 1-Att_*$.}
\vspace{-0.35cm}
\label{fig:overall}
\end{figure*}

In this paper, we propose a novel CoNet for RGB-D SOD.
The overall architecture is shown in Fig.~\ref{fig:overall}.
In this framework, three mutually beneficial collaborators, namely edge detection, coarse salient object detection and depth estimation, work together to aid accurate SOD through exploratory learning and timely interaction.
From different perspectives of the SOD target, knowledges from edge, depth and saliency are fully exploited in a mutual-benefit manner to enhance the detector's performance.
A simplified workflow is given below.

First, a backbone network is used to extract features from original images.
Five transition layers and a global guidance module (GGM) are followed to perform feature preprocessing and generate the integrated low-level feature $f_l$ and high-level feature $f_h$ (details are shown in Sec.~\ref{subsection:3.2}).
Then an edge collaborator is assigned to $f_l$ to extract edge information from the overabundant low-level feature.
For the high-level feature $f_h$, saliency collaborator and depth collaborator work together to jointly enhance the high-level feature learning process of global semantics in a mutual-benefit manner.
Finally, all learned knowledges from three collaborators ($Att_{edge}$, $Att_{sal}$ and $Att_{depth}$), as well as the integrated low-level and high-level feature ($F_g$), are uniformly handed to a knowledge collector (KC).
Here, acting as a tutor, KC summarizes the learned edge, depth and saliency knowledges and utilizes them to predict accurate saliency results.
We elaborate on the three collaborators and the KC in Sec.~\ref{subsection:3.3}.

\begin{table}[t]
	\centering
	\vspace{0.45cm}
	\caption{Detailed information of the five transition layers in Fig.~\ref{fig:overall}. We show the input size and output size of the feature maps before and after those transition layers, and represent their specific transition operators for better understanding.}
	\vspace{-0.15cm}
	\resizebox{!}{1.01cm}{
	\begin{tabular}{c|c|c|c}
		\hline
		\multirow{1}{*}{Transition}           & \multicolumn{1}{|c|}{Input Size} 		&\multicolumn{1}{|c|}{Transition Operators}              &\multicolumn{1}{|c}{Output Size}	\\ \cline{1-4} 
		trans1		&$128\times 128\times 64$      		&$Upsample_{\times 2}$ 	&$256\times 256\times 64$ \\ 
		trans2		&$64\times 64\times 256$			&$Upsample_{\times 4}$	&$256\times 256\times 256$\\ 
		trans3		&$32\times 32\times 512$			&$Upsample_{\times 2}, Conv_{3\times 3}+BN+PRelu$	&$64\times 64\times 64$\\ 
		trans4		&$16\times 16\times 1024$		&$Upsample_{\times 4}, Conv_{3\times 3}+BN+PRelu$	&$64\times 64\times 64$\\ 
		trans5		&$16\times 16\times 2048$		&$Upsample_{\times 4}, Conv_{3\times 3}+BN+PRelu$	&$64\times 64\times 64$\\ \hline
	\end{tabular}}
		\vspace{-0.35cm}
	\label{tab:transition}
\end{table}

\subsection{Feature Preprocessing}
\label{subsection:3.2}
\paragraph{\bf Backbone Network.}
We use the widely used ResNet~\cite{ResNet} suggested by other deep-learning-based methods~\cite{2DR3Net,2DPiCANet,2DCPD} as backbone network, where the last fully connected layers are truncated to better fit for the SOD task.
As shown in Fig.~\ref{fig:overall}, five side-out features generated from the backbone network are transferred to five transition layers to change their sizes and the number of channels.
Detailed parameters are listed in Table.~\ref{tab:transition}, and the five output features are defined as $\{f_1, f_2, f_3, f_4, f_5\}$.
\\
\begin{wrapfigure}{r}{0.5\textwidth}
\vspace{-0.05cm}
\centering 
\includegraphics [width=0.79\linewidth] {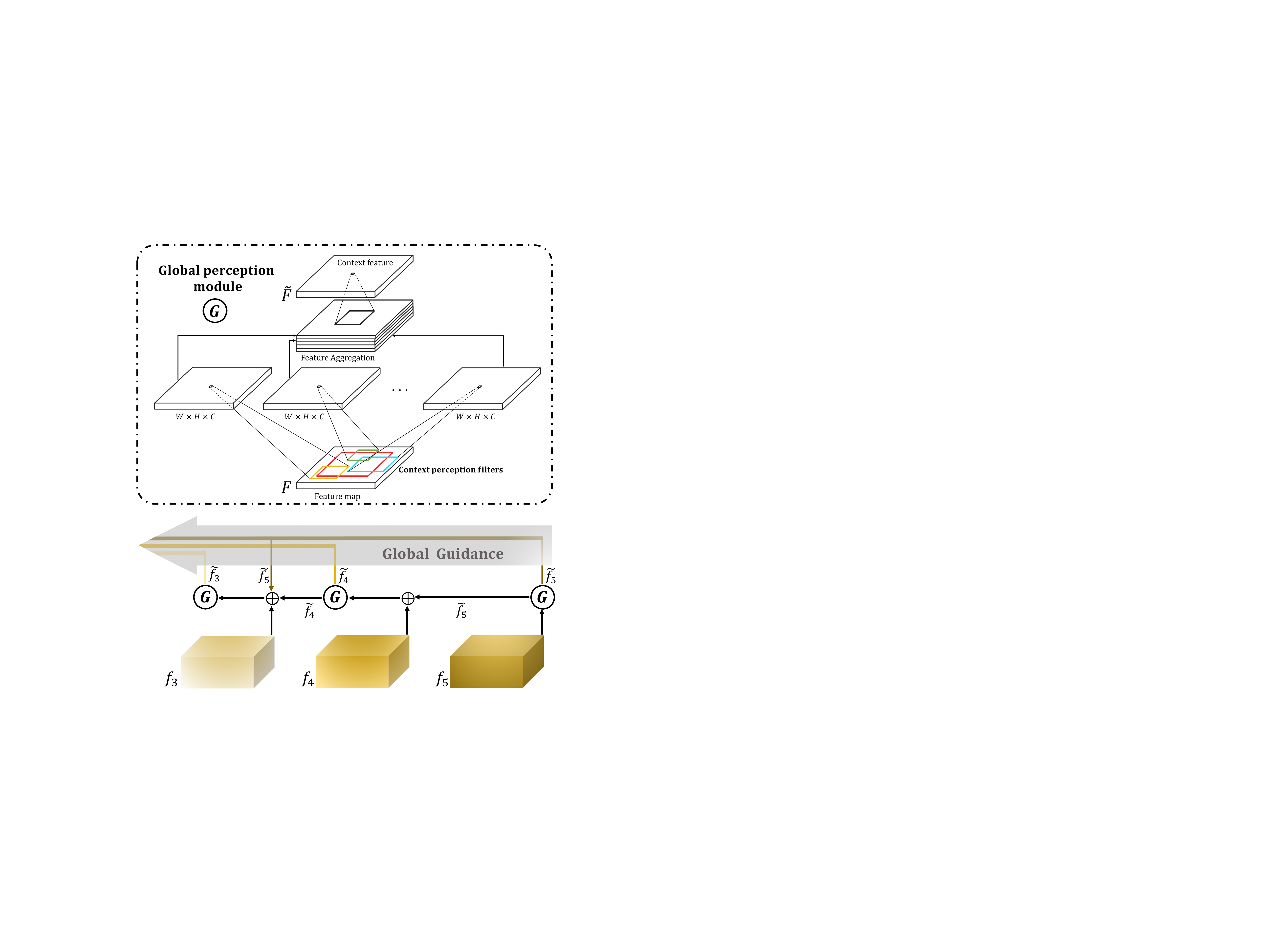}
\vspace{-0.25cm}
\caption{The architecture of global guidance module (GGM).}
\vspace{-0.55cm}
\label{fig:GGM}
\end{wrapfigure}
\noindent{\bf Global Guidance Module.}
In order to obtain richer global semantics and alleviate information dilution in the decoder, a Global Guidance Module (GGM) is applied on high-level features (i.e. $f_3, f_4$, and $f_5$)(see Fig.~\ref{fig:GGM}).
Its key component, global perception module (GPM), takes the progressively integrated feature as input, followed by four parallel dilated convolution operations~\cite{Dilation} (kernel size = 3, dilation rates = 1/6/12/18 ) and one 1$\times$1 traditional convolution operation, to obtain rich global semantics.
Benefiting from the dilated convolution~\cite{Dilation}, the GPM captures affluent multi-scale contextual information without sacrificing image resolution~\cite{DeepLab1,DeepLab3}.
Here, we define the process of GPM as $\widetilde{F} = \Phi(F)$, where $F$ denotes the input feature map and $\widetilde{F}$ means the output feature.
In GGM, we take the summation of the feature in current layer and the output features of all high-level GPMs as input to alleviate information dilution.
Finally, three output features of GPMs are concatenated and an integrated high-level feature $f_h$ is produced, which is computed by:
\begin{equation}
\begin{matrix} \widetilde{f}_i= \Phi(f_i+\sum_{m=i+1}^{5}\widetilde{f}_m), i = 3,4,5, \end{matrix}
\end{equation}
\begin{equation}
f_h=Up(W_h*Concat(\widetilde{f}_3, \widetilde{f}_4, \widetilde{f}_5) + b_h),
\end{equation}
where * means convolution operation. $W_h$ and $b_h$ are convolution parameters. $Up(\cdot)$ means the upsampling operation.

\subsection{Collaborative Learning}
\label{subsection:3.3}
\noindent{\bf Edge Collaborator.}
Existing 3D methods~\cite{3DDF,3DPCA,3DTANet,3DDMRA,3DCPFP} have achieved remarkable performance in locating salient regions, but they still suffer from coarse object boundaries.
In our framework, we design an edge collaborator to explicitly extract edge information from the overabundant low-level feature and use this information to give more emphasis to object boundaries.

Specifically, we first formulate this problem by adding edge supervision on the top of integrated low-level feature $f_l$.
The used edge ground truths (GT) (shown in Fig.~\ref{fig:introduction}) are derived from saliency GT using canny operator~\cite{Canny}.
As shown in Fig.~\ref{fig:overall}, $f_l$ is processed by a 1$\times$1 convolution operation and a softmax function to generate the edge map $M_{edge}$.
Then, binary cross entropy loss (denoted as ${loss}_{e}$) is adopted to calculate the difference between $M_{edge}$ and edge GT.
As the edge maps $M_{edge}$ in Fig.~\ref{fig:overall} and Fig.~\ref{fig:tri-att} show,
edge detection constraint is beneficial for predicting accurate boundaries of salient objects.
Additionally, we also transfer the learned edge knowledge before the softmax function (denoted as $Att_{edge}$) to the knowledge collector (KC), where the edge information is further utilized to emphasize object boundaries.
The reason why we use $Att_{edge}$ rather than $M_{edge}$ is to alleviate the negative influence brought by accuracy decrement of $M_{edge}$.

\noindent{\bf Saliency and Depth Collaborators.}
When addressing scene understanding tasks like semantic segmentation and salient object detection, there exist strong consistencies among the global semantics and geometric properties of image regions~\cite{Towards}.
In our framework, a saliency collaborator and a depth collaborator work together to jointly enhance the feature learning process of high-level semantics in a mutual-benefit manner.

\emph{ Stage one:}
The high-level feature $f_h$ is first processed by a 1 $\times$ 1 convolution operation and a softmax function to predict a coarse saliency map $S_{coarse}$.
Here, binary cross entropy loss (denoted as $loss_{s}$) is used for training.
Then, the learned saliency knowledge acts as a spatial attention map to refine the high-level feature in a similar way like~\cite{2DCPD}.
But different from~\cite{2DCPD} which considers $S_{coarse}$ as attention map directly, we use the more informative feature map before softmax function (denoted as $Att_{sal}$) to emphasize or suppress each pixel of $f_h$ .
Identify mapping is adopted to alleviate the errors in $Att_{sal}$ to be propagated to depth learning and accelerate network convergence.
Formally, this procedure can be defined as:
\begin{equation}
Att_{sal}= W_s*f_h+b_s,
\end{equation}
\begin{equation}
\widetilde{f}_h = Att_{sal}\odot f_h + f_h,
\end{equation}
where $\odot$ means element-wise multiplication. $\widetilde{f}_h$ denotes the output saliency-enhanced feature.

\emph{ Stage two:} 
As pointed out in previous RGB-D researches~\cite{3DDMRA,3DCPFP}, the spatial information within depth image is helpful for better locating salient objects in a scene.
In our network, we innovatively integrate depth learning into the high-level feature learning process, instead of directly taking depth image as input.
This learning strategy enables our network to be free of using an extra depth network to make inference from an extra depth input,
and thus being more lightweight and versatile.
As in Fig.~\ref{fig:overall}, a depth head with three convolution layers (defined as $\Psi(\cdot)$) is first used to make feature $\widetilde{f}_h$ adapt to depth estimation.
Then, its output $\Psi(\widetilde{f}_h)$ is followed by a 1 $\times$ 1 convolution operation to generate the estimated depth map $Att_{depth}$.
Here, depth images act as GTs for supervision and we use smooth $L_{1}$ loss~\cite{FastRCNN} to calculate the difference between $Att_{depth}$ and depth GT,
where smooth $L_1$ loss is a robust $L_1$ loss proposed in~\cite{FastRCNN} that is less sensitive to outliers than $L_2$ loss.
Formally, the depth loss can be defined as:
\begin{equation}
\begin{small}
Loss_{d} = \frac{1}{W\times H}\sum_{x=1}^W \sum_{y=1}^H\\
\begin{cases} 
0.5\times|\triangle(x,y)|^2, & \text{if }|\triangle(x,y)|\le1, \\
|\triangle(x,y)|-0.5, & \text{if }\triangle(x,y)<-1 \text{ or } \triangle(x,y)>1,
\end{cases}
\end{small}
\end{equation}
where $W$ and $H$ denote the width and height of the depth map. $\triangle(x,y)$ means the error between prediction $Att_{depth}$ and the depth GT in each pixel $(x,y)$.
Since each channel of a feature map can be considered as a ‘feature detector’~\cite{CBAM}, the depth knowledge $Att_{depth}$ is further employed to learn a channel-wise attention map $M_c$ for choosing useful semantics.
Identify mapping operation is also adopted to enhance the fault-tolerant ability.
This procedure can be defined as: 
\begin{equation}
Att_{depth} = W_d*\Psi(\widetilde{f}_h) + b_d,
\end{equation}
\begin{equation}
M_c = \sigma(GP(W_c*Att_{depth}+b_c)),
\end{equation}
\begin{equation}
f_{hc} = M_c\otimes \widetilde{f}_h+\widetilde{f}_h,
\end{equation}
where $w_*$ and $b_*$ are parameters to be learned. $GP(\cdot)$ means global pooling operation. $\sigma(\cdot)$ is the softmax function. $\otimes$ denotes channel-wise multiplication.

After these two stages, two collaborators can cooperatively generate optimal feature which contains affluent spatial cues and possesses strong ability to distinguish salient and non-salient regions.

\noindent{\bf Knowledge Collector.}
\label{subsection:3.4}
In our framework, the KC works as a unified tutor to complete knowledge transfer from individual collaborators to the group.

As illustrated in Fig.~\ref{fig:overall}, all knowledges learned from three collaborators (i.e. $Att_{edge}$, $Att_{sal}$, and $Att_{depth}$) and the concatenated multi-level feature $F_g=Concat(f_l,f_{hc})$ are uniformly transferred to the KC. 
Those information are comprehensively processed in a triple-attention manner to give more emphasis to salient regions and object boundaries.
In Fig.~\ref{fig:overall}, we show a detailed diagram with visualized attention maps for better understanding.
To be specific, $Att_{edge}$ and $Att_{sal}$ are first concatenated together to jointly learn a fused attention map $Att_f$, where the locations and boundaries of the salient objects are considered uniformly.
Then, $F_g$  is in turn multiplied with the depth attention map $Att_{depth}$ and the fused attention map $Att_f$, which significantly enhances the contrast between salient and non-salient areas.
Ablation analysis shows the ability of the KC to enhance the performance significantly.

There is a vital problem worth thinking about.
The quality of $Att_{depth}$ and $Att_f$ might lead to irrecoverable inhibition of salient areas.
Therefore, we add several residual connection operations~\cite{ResNet} to the KC to retain the original features.
Formally, this process can be defined as:
\begin{equation}
Att_{f} = \sigma(W_f*Concat(Att_{sal}, Att_{edge})+b_f),
\end{equation}
\begin{equation}
\widetilde{F}_g = Att_{depth}\odot F_g+F_g,
\end{equation}
\begin{equation}
F = Att_f\odot \widetilde{F}_g+\widetilde{F}_g.
\end{equation}
In the end, $F$ is followed by a 1 $\times$ 1 convolution operation and an upsampling operation to generate the final saliency map $S_{final}$.
Here, binary cross entropy loss (denoted as $loss_{f}$) is used to calculate the difference between $S_{final}$ and saliency GT.
Thus, the total loss $L$ can be represented as:
\begin{equation}
\begin{small}
L = \lambda_eLoss_{e}+\lambda_sLoss_{s}+\lambda_dLoss_{d}+\lambda_fLoss_{f},
\end{small}
\end{equation}
where $Loss_e$, $Loss_s$, and $Loss_f$ are cross entropy loss and $Loss_d$ is a smooth $L_1$ loss.
In this paper, we set $\lambda_e=\lambda_s=\lambda_f=1$ and $\lambda_d=3$.

\section{Experiments}

\subsection{Dataset}
To evaluate the performance of our network, we conduct experiments on seven widely used benchmark datasets.

\noindent{\bf DUT-D~\cite{3DDMRA}:} contains 1200 images with 800 indoor and 400 outdoor scenes paired with corresponding depth images. This dataset contains many complex scenarios.
\noindent{\bf NJUD~\cite{3DACSD}:} contains 1985 stereo images (the latest version). They are gathered from the Internet, 3D movies and photographs taken by a Fuji W3 stereo camera.
\noindent{\bf NLPR~\cite{3DNLPR}:} includes 1000 images captured by Kinect under different illumination conditions.
\noindent{\bf SIP~\cite{3DSIP}:} contains 929 salient person samples with different poses and illumination conditions.
\noindent{\bf LFSD~\cite{LFS2017}:} is a relatively small dataset with 100 images captured by Lytro camera.
\noindent{\bf STEREO~\cite{3DSTEREO}:} contains 797 stereoscopic images downloaded from the Internet.
 \noindent{\bf RGBD135~\cite{3DDES}:} consists of seven indoor scenes and contains 135 images captured by Kinect.

For training, we split 800 samples from DUT-D, 1485 samples from NJUD, and 700 samples from NLPR as in~\cite{3DPCA,3DTANet,3DDMRA}.
The remaining images and other public datasets are all for testing to comprehensively evaluate the generation abilities of models.
To reduce overfitting, we augment the training set by randomly flipping, cropping and rotating those images.

\subsection{Experimental setup}
\noindent{\bf Evaluation metrics.} 
We adopt 6 widely used evaluation metrics to verify the performance of various models, including precision-recall (PR) curve, mean F-measure ($F_{\beta}$)~\cite{Fmeasure}, mean absolute error (MAE)~\cite{Benchmark}, weighted F-measure ($F_{\beta}^w$)~\cite{weightedF} and recently proposed S-measure ($S$)~\cite{Smeasure} and E-measure ($E$)~\cite{Emeasure}.
Saliency maps are binarized using a series of thresholds and then pairs of precision and recall are computed to plot the PR curve.
The F-measure is a harmonic mean of average precision and average recall.
Here, we calculate the mean F-measure which uses adaptive threshold to generate binary saliency map.
The MAE represents the average absolute difference between the saliency map and ground truth.
Weighted F-measure intuitively generalizes F-measure by alternating the way to calculate the Precision and Recall.
S-measure contains two terms: object-aware and region-aware structural similarities.
E-measure jointly captures image level statistics and local pixel matching information.
Details of those evaluation metrics can refer to~\cite{2DDeep}.
For MAE, lower value is better. For others, higher is better.

\noindent{\bf Implementation details.} 
We implement our proposed framework using the Pytorch toolbox and train it with a GTX 1080 Ti GPU.
All training and test images are uniformly resized to $256\times256$.
Our network is trained in an end-to-end manner using the standard SGD optimizer, and it converges after 50 epochs with batch size of 2.
The momentum, weight decay and learning rate are set as 0.9, 0.0005 and 1e-10, respectively.
Any post-processing procedure (e.g., CRF~\cite{CRF}) is not applied in this work.
The model size of our network has only 167.6M and the inference speed for a 256 $\times$ 256 image only takes 0.0290s (34FPS).

\subsection{Ablation Analysis}
\begin{wrapfigure}{r}{0.5\textwidth}
\begin{minipage}[t]{0.5\textwidth}
\begin{minipage}[t]{1\textwidth}
  \centering
  \vspace{-0.9cm}
     \makeatletter\def\@captype{table}\makeatother\caption{Quantitative results of the ablation analysis on two benchmark datasets. B means the backbone network. E and S represent edge supervision and saliency supervision respectively. $S_{SA}+D_{CA}$ means our mutual-benefit learning strategy between depth and saliency. +KC means adding our knowledge collector on (e).}
     \vspace{0.2cm}
     \resizebox{!}{1.42cm}{
      \begin{tabular}{clccccccccccccc}
		\toprule
		\multicolumn{1}{c}{}	&\multicolumn{1}{c}{} 	& \multicolumn{2}{c}{NJUD} 							&\multicolumn{2}{c}{NLPR}              					\\ 
								\cmidrule(r){3-4}								 \cmidrule(r){5-6}
		indexes	&Modules		&$F_{\beta}\uparrow$	&$MAE\downarrow$	&$F_{\beta}\uparrow$	&$MAE\downarrow$	  \\ 
			\midrule
			(a)		&B						&0.831	&0.065	&0.797	&0.050\\
			(b)		&B+GGM					&0.839	&0.060	&0.813	&0.044\\
			(c)		&(b)+E					&0.851	&0.056	&0.825	&0.041\\
			(d)		&(b)+E+S					&0.857	&0.054	&0.833	&0.038\\
			(e)		&(b)+E+$S_{SA}$+$D_{CA}$	&0.864	&0.051	&0.841	&0.035\\
			(f)		&(e)+KC					&\textbf{0.872}	&\textbf{0.047}	&\textbf{0.848}&\textbf{0.031}\\		
		\bottomrule
	\end{tabular}}
	\label{tab:ablation}
  \end{minipage}
  \vfill
\begin{minipage}[t]{1\linewidth}
\vspace{0.45cm}
  \centerline{\includegraphics[width=6.1cm]{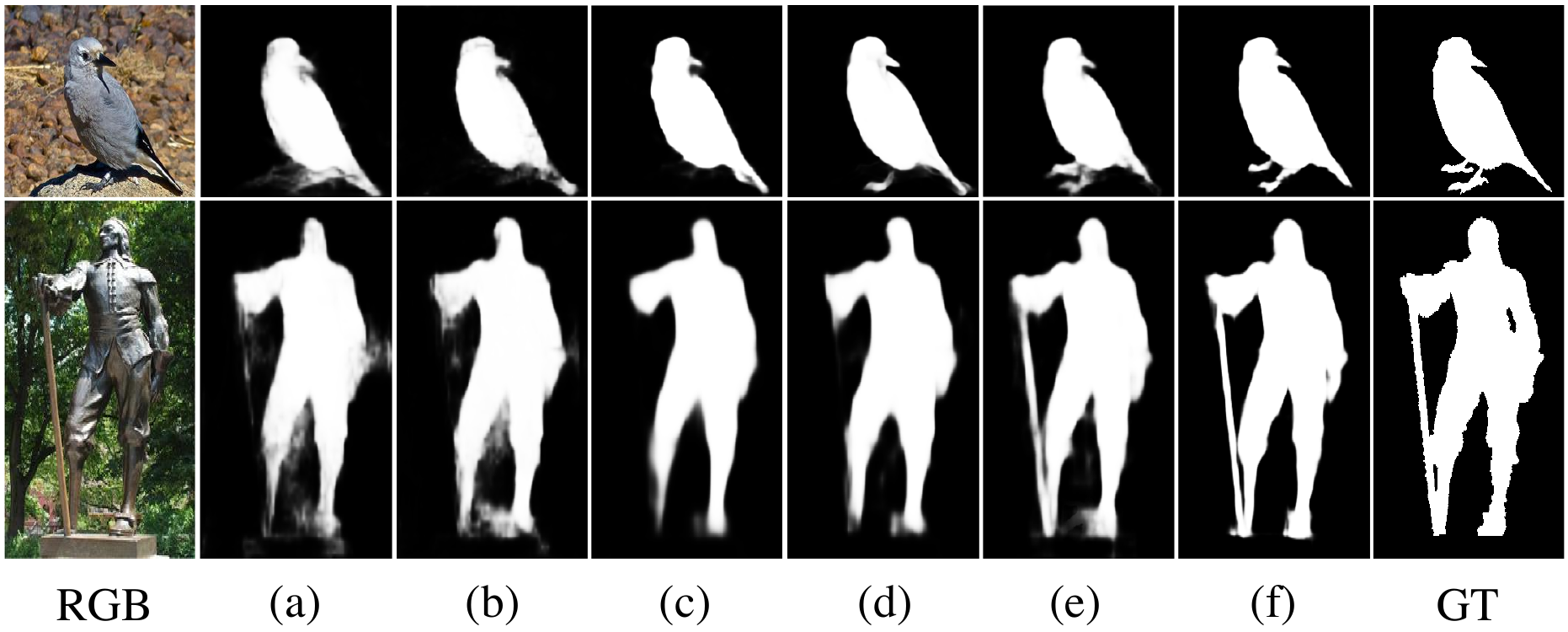}}
  \vspace{-0.3cm}
  \caption{Visual saliency maps of ablation analysis. The meaning of the indexes (a)-(f) can refer to Table.~\ref{tab:ablation}.}
  \vspace{-0.4cm}
  \label{fig:ablation}
\end{minipage}
\end{minipage}
\label{fig:res}
\end{wrapfigure}

\begin{wrapfigure}{r}{0.52\textwidth}
\begin{minipage}[t]{0.52\textwidth}
\vspace{0.15cm}
	\begin{minipage}[t]{1\textwidth}
  \centering
  \vspace{-1.1cm}
     \makeatletter\def\@captype{table}\makeatother\caption{Ablation analysis of the interactions between three collaborators. The meaning of indexes (b)-(f) can refer to Table.~\ref{tab:ablation}. $+S_l$ means adding saliency supervision on low-level feature. D means depth supervision.}
	\vspace{0.25cm}
	\resizebox{!}{2.82 cm}{
	\begin{tabular}{lcccc}
		\toprule
		\multicolumn{1}{c}{}    & \multicolumn{2}{c}{NJUD} 							&\multicolumn{2}{c}{NLPR}   \\ 
								\cmidrule(r){2-3}		 \cmidrule(r){4-5}
			Modules			&$F_{\beta}\uparrow$	&$MAE\downarrow$ 	&$F_{\beta}\uparrow$&	 $MAE\downarrow$\\ \midrule \midrule
			\multicolumn{5}{l}{\textbf{Saliency $\&$ Edge}} \\ 
						(b)				&0.839	&0.060	&0.813	&0.044\\	
						(b)+E (c)					&0.851	&0.056	&0.825	&0.041\\	
						(b)+E+$S_l$				&0.835	&0.062	&0.807	&0.044\\\midrule\midrule
			\multicolumn{5}{l}{\textbf{Saliency $\&$ Depth}}\\
						(c)						&0.851	&0.056	&0.825	&0.041\\	
						(c)+$S$ (d)				&0.857	&0.054	&0.833	&0.038\\
						(c)+$S$+D				&0.859     &0.054	&0.835	&0.037\\
						(c)+$S$+$D_{CA}$			&0.861	&0.053	&0.837	&0.036\\
						(c)+$S_{SA}$+$D_{CA}$ (e)	&0.864	&0.051	&0.841	&0.035\\\midrule\midrule
			\multicolumn{5}{l}{\textbf{Saliency $\&$ Edge $\&$ Depth}} \\
						(e)							&0.864	&0.051	&0.841	&0.035\\
						(e)+$Att_{edge}$				&0.868	&0.049	&0.846	&0.032\\
						(e)+$Att_{sal}$					&0.866	&0.049	&0.844	&0.033\\
				(e)+$Att_{edge}$+$Att_{sal}$			       &0.869	&0.048	&0.846	&0.031\\
                    (e)+$Att_{edge}$+$Att_{sal}$+$Att_{depth}$ (f)	 	&0.872	&0.047	&0.848	&0.031\\\bottomrule
	\end{tabular}}
	\label{tab:mutual_ablation}
 	 \end{minipage}
 	 \vfill
	\begin{minipage}[t]{1\linewidth}
\vspace{0.2cm}
  \centerline{\includegraphics[width=5.99cm]{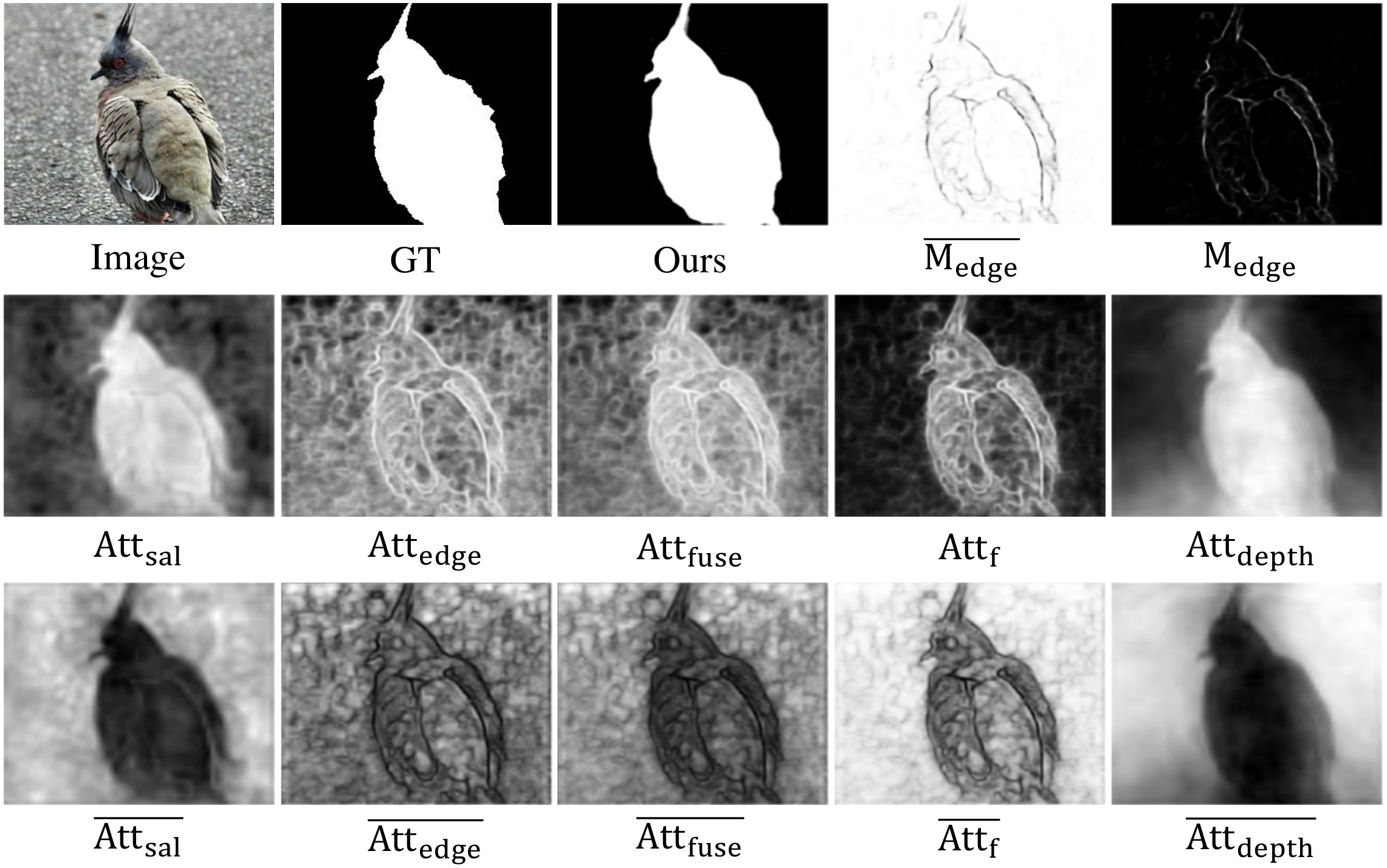}}
  \vspace{-0.3cm}
  \caption{Internal results in the knowledge collector. The results of another sample can be seen in Fig.~\ref{fig:overall}. Here, $\overline{F} = 1-F$.}
  \label{fig:tri-att}
\end{minipage}
\end{minipage}
\vspace{-0.55cm}
\label{fig:res}
\end{wrapfigure}

\noindent{\bf Overview of Performance.}
We show the quantitative and qualitative results of different modules of our proposed network in Tab.~\ref{tab:ablation} and Fig.~\ref{fig:ablation}.
The backbone network (denoted as $B$) is constructed by directly concatenating low-level feature $f_l$ and high-level feature $f_h$ without using GGM for prediction.
Comparison of the results (a) and (b) shows that adding our GGM can more effectively extract rich semantic features and prevent information dilution in the decoding stage.
After introducing edge supervision (denoted as E), the boundaries of the saliency maps are sharper (b $vs$ c in Fig.~\ref{fig:ablation}).
The edge maps ($M_{edge}$) in Fig.~\ref{fig:overall} and Fig.~\ref{fig:tri-att} also show the ability of our network in explicitly extracting object boundaries.
By adding additional saliency supervision on $f_h$ (denoted as S), the performance can be further improved.
However, by comparing (d) and (e), we can see that our mutual-benefit learning style between saliency collaborator and depth collaborator (denoted as $S_{SA}$ and $D_{CA}$) can further improve the detector's ability to locate salient objects.
This also verifies the strong correlation between saliency and depth.
Finally, by using our proposed (KC), all learned edge, depth and saliency knowledges from three collaborators can be effectively summarized and utilized to give more emphasis to salient regions and object boundaries, improving the average MAE performance on two datasets by nearly 9.6\% points.
By comparing (e) and (f) in Fig.~\ref{fig:ablation}, we can also see that salient regions in (f) are more consistent with the saliency GT and the object boundaries are explicitly highlighted benefiting from the comprehensive knowledge utilization.
Those advances demonstrate that using our collaborative learning strategy is beneficial for accurate saliency prediction.
We list some numerical results here for better understanding. The Root Mean Squared Error (RMSE) of depth prediction on NJUD and NLPR datasets are 0.3684 and 0.4696, respectively.
The MAE scores of edge prediction are 0.053 and 0.044, respectively.

\noindent{\bf The Interactions between Collaborators.}\\
\noindent{\bf\emph{Saliency and Edge.}}
To explore the correlation between saliency and edge, we gradually add edge detection supervision (denoted as $E$) and saliency supervision (denoted as $S_l$) on the low-level feature $f_l$.
From the quantitive results in Tab.~\ref{tab:mutual_ablation}, we can see that adding edge supervision can explicitly extract clear boundary information and significantly enhance the detection performance, especially for the F-measure scores.
However, when adding saliency supervision on $f_l$, the performances on both datasets decrease dramatically.
This is partly because the low-level features contain too much information and are relatively too coarse to predict saliency, and partly because the two tasks are to some extent incompatible, in which one is for highlighting the boundaries and another is for highlighting the whole salient objects.
Hence, it is optimal to only add edge detection supervision on the low-level feature.
\\
\noindent{\bf\emph{ Saliency and Depth.}}
In order to verify the effectiveness of the proposed mutual-benefit learning strategy on high-level feature $f_h$, we gradually add two collaborators and their mutual-benefit operations to the baseline model (c).
As shown in Tab.~\ref{tab:mutual_ablation}, adding saliency supervision (denoted as S) and adding depth supervision (denoted as D) are all beneficial for extracting more representative high-level semantic features.
In addition, by gradually introducing our proposed mutual-benefit learning strategy between two collaborators (denoted as $S_{SA}$ and $D_{CA}$), spatial layouts and global semantics of high-level feature can be greatly enhanced, which consequently brings additional accuracy gains on both datasets.
These results further verify the effectiveness of our collaborative learning framework.
\\
\noindent{\bf\emph{ Saliency, Edge and Depth.}}
In our knowledge collector, all knowledge learned from three collaborators are summarized and utilized in a triple-attention manner.
As the visualized attention maps in Fig.~\ref{fig:overall} and Fig.~\ref{fig:tri-att} show, the edge knowledge ($Att_{edge}$) can help highlight object boundaries, and the depth and saliency knowledge ($Att_{depth}$ and $Att_{sal}$) can also be used to emphasize salient regions and suppress non-salient regions.
We can see from Tab.~\ref{tab:mutual_ablation} that both $Att_{edge}$ and $Att_{sal}$ are beneficial for enhancing the feature representation and improving the F-measure and MAE performance.
In our framework, we adopt a better strategy that $Att_{edge}$ and $Att_{sal}$ are concatenated together to jointly emphasize salient objects and their boundaries.
Finally, by comparing the results in the last two lines of Tab.~\ref{tab:mutual_ablation}, we can see that by further utilizing the learned depth knowledge, the detector's performance can be further improved.
We visualize all internal results of the KC in Fig.~\ref{fig:tri-att} for better understanding.

\begin{table*}[tp]
	\caption{Quantitative comparisons on seven benchmark datasets. The best three results are shown in \textbf{\textcolor{blue}{blue}}, \textcolor{red}{red}, and \textcolor{green}{green} fonts respectively.}
	\vspace{-0.25cm}
	\centering
	\resizebox{!}{5.2cm}{
	\begin{tabular}{lllccccccccccccc}
		\toprule
		 \textbf{Dataset} 	&\textbf{Metric} &\textbf{DES}	&\textbf{LHM} &\textbf{DCMC} 	&\textbf{MB} 	&\textbf{CDCP} &\textbf{DF} 	&\textbf{CTMF}  &\textbf{PDNet} &\textbf{MPCI} &\textbf{TANet} &\textbf{PCA} &\textbf{CPFP} &\textbf{DMRA} &\textbf{Ours}\\
		 	      			&			&\cite{3DDES}	&\cite{3DNLPR}&\cite{3DDCMC}	&\cite{3DMB}	&\cite{3DCDCP}&\cite{3DDF}	&\cite{3DCTMF}&\cite{3DPDNet}&\cite{3DMPCI}&\cite{3DTANet}&\cite{3DPCA}&\cite{3DCPFP}&\cite{3DDMRA}&\\
			\midrule\midrule
			
	\multirow{5}{*}{\textbf{DUT-D}~\cite{3DDMRA}}
&$E\uparrow$		&0.733	&0.767	&0.712	&0.691	&0.794	&0.842	&\textcolor{green}{0.884}	&0.861	&0.855	&0.866	&0.858	&0.854	&\textcolor{red}{0.927}	&\textbf{\textcolor{blue}{0.941}}\\
&$S\uparrow$	&0.659	&0.568	&0.499	&0.607	&0.687	&0.730	&\textcolor{green}{0.834}	&0.799	&0.791	&0.808	&0.801	&0.749	&\textcolor{red}{0.888}	&\textbf{\textcolor{blue}{0.918}}\\
&$F_{\beta}^w\uparrow$				&0.386	&0.350	&0.290	&0.464	&0.530	&0.542	&0.690	&0.650	&0.636	&\textcolor{green}{0.712}	&0.696	&0.644	&\textcolor{red}{0.858}	&\textbf{\textcolor{blue}{0.896}}\\
&$F_{\beta}\uparrow$	&0.668	&0.659	&0.406	&0.577	&0.633	&0.748	&\textcolor{green}{0.792}	&0.757	&0.753	&0.779	&0.760	&0.736	&\textcolor{red}{0.883}	&\textbf{\textcolor{blue}{0.908}}\\
&$MAE\downarrow$		&0.280	&0.174	&0.243	&0.156	&0.159	&0.145	&0.097	&0.112	&0.113	&\textcolor{green}{0.093}	&0.100	&0.099	&\textcolor{red}{0.048}	&\textbf{\textcolor{blue}{0.034}}\\
									\cmidrule(r){1-16}
									
	\multirow{5}{*}{\textbf{NJUD}~\cite{3DACSD}}
&$E\uparrow$		&0.421	&0.722	&0.796	&0.643	&0.751	&0.818	&0.864	&0.890	&0.878	&0.893	&\textcolor{green}{0.896}	&0.894	&\textcolor{red}{0.908}	&\textbf{\textcolor{blue}{0.912}}\\
&$S\uparrow$	&0.413	&0.530	&0.703	&0.534	&0.673	&0.735	&0.849	&\textcolor{green}{0.883}	&0.859	&0.878	&0.877	&0.878	&\textcolor{red}{0.886}	&\textbf{\textcolor{blue}{0.894}}\\
&$F_{\beta}^w\uparrow$				&0.241	&0.311	&0.506	&0.369	&0.522	&0.552	&0.732	&0.798	&0.749	&0.812	&0.811	&\textcolor{green}{0.837}	&\textcolor{red}{0.853}	&\textbf{\textcolor{blue}{0.856}}\\
&$F_{\beta}\uparrow$	&0.165	&0.625	&0.715	&0.492	&0.618	&0.744	&0.788	&0.832	&0.813	&\textcolor{green}{0.844}	&\textcolor{green}{0.844}	&\textcolor{red}{0.850}	&\textbf{\textcolor{blue}{0.872}}	&\textbf{\textcolor{blue}{0.872}}\\
&$MAE\downarrow$		&0.448	&0.201	&0.167	&0.202	&0.181	&0.151	&0.085	&0.062	&0.079	&0.061	&0.059	&\textcolor{green}{0.053}	&\textcolor{red}{0.051}	&\textbf{\textcolor{blue}{0.047}}\\
										\cmidrule(r){1-16}
	\multirow{5}{*}{\textbf{NLPR}~\cite{3DNLPR}}
&$E\uparrow$		&0.735	&0.772	&0.684	&0.814	&0.785	&0.838	&0.869	&0.876	&0.871	&0.916	&0.916	&\textcolor{green}{0.924}	&\textbf{\textcolor{blue}{0.942}}	&\textcolor{red}{0.936}\\
&$S\uparrow$	&0.582	&0.591	&0.550	&0.714	&0.724	&0.769	&0.860	&0.835	&0.855	&0.886	&0.873	&\textcolor{green}{0.888}	&\textcolor{red}{0.899}	&\textbf{\textcolor{blue}{0.907}}\\
&$F_{\beta}^w\uparrow$			&0.259	&0.320	&0.265	&0.574	&0.512	&0.524	&0.691	&0.659	&0.688	&0.789	&0.772	&\textcolor{green}{0.820}	&\textcolor{red}{0.845}	&\textbf{\textcolor{blue}{0.850}}\\
&$F_{\beta}\uparrow$	&0.583	&0.520	&0.328	&0.637	&0.591	&0.682	&0.723	&0.740	&0.729	&0.795	&0.794	&\textcolor{green}{0.822}	&\textbf{\textcolor{blue}{0.855}}	&\textcolor{red}{0.848}\\
&$MAE\downarrow$		&0.301	&0.119	&0.196	&0.089	&0.114	&0.099	&0.056	&0.064	&0.059	&\textcolor{green}{0.041}	&0.044	&\textcolor{red}{0.036}	&\textbf{\textcolor{blue}{0.031}}	&\textbf{\textcolor{blue}{0.031}}	\\

									\cmidrule(r){1-16}
									
	\multirow{5}{*}{\textbf{STEREO}~\cite{3DSTEREO}} 
&$E\uparrow$		&0.451	&0.781	&0.838	&0.693	&0.801	&0.844	&0.870	&0.903	&0.890	&\textcolor{green}{0.911}	&0.905	&0.897	&\textcolor{red}{0.920}	&\textbf{\textcolor{blue}{0.923}}\\
&$S\uparrow$	&0.473	&0.567	&0.745	&0.579	&0.727	&0.763	&0.853	&0.874	&0.856	&0.877	&\textcolor{green}{0.880}	&0.871	&\textcolor{red}{0.886}	&\textbf{\textcolor{blue}{0.908}}\\
&$F_{\beta}^w\uparrow$			&0.277	&0.369	&0.551	&0.445	&0.595	&0.576	&0.727	&0.799	&0.747	&0.811	&0.810	&\textcolor{green}{0.818}	&\textcolor{red}{0.850}	&\textbf{\textcolor{blue}{0.871}}\\
&$F_{\beta}\uparrow$	&0.223	&0.716	&0.761	&0.572	&0.680	&0.761	&0.786	&0.833	&0.812	&\textcolor{green}{0.849}	&0.845	&0.827	&\textcolor{red}{0.868}	&\textbf{\textcolor{blue}{0.885}}\\
&$MAE\downarrow$			&0.417	&0.179	&0.150	&0.178	&0.149	&0.142	&0.087	&0.064	&0.080	&0.060	&0.061	&\textcolor{green}{0.054}	&\textcolor{red}{0.047}	&\textbf{\textcolor{blue}{0.041}}\\

									\cmidrule(r){1-16}
	\multirow{5}{*}{\textbf{SIP}~\cite{3DSIP}}
&$E\uparrow$		&0.742	&0.722	&0.787	&0.715	&0.721	&0.794	&0.824	&0.802	&0.886	&0.893	&\textcolor{green}{0.898}	&\textcolor{red}{0.899}	&0.863	&\textbf{\textcolor{blue}{0.909}}\\
&$S\uparrow$	&0.616	&0.523	&0.684	&0.624	&0.597	&0.651	&0.716	&0.691	&0.833	&0.835	&\textcolor{green}{0.844}	&\textcolor{red}{0.850}	&0.806	&\textbf{\textcolor{blue}{0.858}}\\
&$F_{\beta}^w\uparrow$			&0.352	&0.286	&0.426	&0.474	&0.411	&0.411	&0.551	&0.503	&0.726	&0.762	&\textcolor{green}{0.777}	&\textcolor{red}{0.798}	&0.750	&\textbf{\textcolor{blue}{0.814}}\\
&$F_{\beta}\uparrow$	&0.646	&0.593	&0.646	&0.573	&0.494	&0.672	&0.684	&0.620	&0.795	&0.809	&\textcolor{red}{0.824}	&0.818	&\textcolor{green}{0.819}	&\textbf{\textcolor{blue}{0.842}}\\
&$MAE\downarrow$			&0.300	&0.182	&0.186	&0.163	&0.224	&0.186	&0.139	&0.166	&0.086	&0.075	&\textcolor{green}{0.071}	&\textcolor{red}{0.064}	&0.085	&\textbf{\textcolor{blue}{0.063}}\\
									\cmidrule(r){1-16}

	\multirow{5}{*}{\textbf{LFSD}~\cite{LFS2017}}
&$E\uparrow$		&0.475	&0.742	&0.842	&0.631	&0.737	&0.841	&0.851	&0.872	&0.840	&0.845	&0.846	&\textcolor{green}{0.867}	&\textbf{\textcolor{blue}{0.899}}	&\textcolor{red}{0.897}\\
&$S\uparrow$	&0.440	&0.558	&0.754	&0.538	&0.658	&0.796	&0.796	&\textcolor{green}{0.845}	&0.787	&0.801	&0.800	&0.828	&\textcolor{red}{0.847}	&\textbf{\textcolor{blue}{0.862}}\\
&$F_{\beta}^w\uparrow$			&0.278	&0.379	&0.605	&0.401	&0.524	&0.645	&0.700	&0.738	&0.668	&0.723	&0.720	&\textcolor{green}{0.779}	&\textcolor{red}{0.814}	&\textbf{\textcolor{blue}{0.819}}\\
&$F_{\beta}\uparrow$	&0.228	&0.708	&0.815	&0.543	&0.634	&0.810	&0.781	&\textcolor{green}{0.824}	&0.779	&0.794	&0.794	&0.813	&\textbf{\textcolor{blue}{0.849}}	&\textcolor{red}{0.848}\\
&$MAE\downarrow$			&0.415	&0.211	&0.155	&0.218	&0.199	&0.142	&0.120	&0.109	&0.132	&0.111	&0.112	&\textcolor{green}{0.088}	&\textcolor{red}{0.075}	&\textbf{\textcolor{blue}{0.071}}\\
									\cmidrule(r){1-16}
	\multirow{5}{*}{\textbf{RGBD135}~\cite{3DDES}}
&$E\uparrow$		&0.786	&0.850	&0.674	&0.798	&0.806	&0.801	&0.907	&0.915	&0.899	&\textcolor{green}{0.916}	&0.909	&\textcolor{red}{0.927}	&\textbf{\textcolor{blue}{0.945}}	&\textbf{\textcolor{blue}{0.945}}\\
&$S\uparrow$	&0.627	&0.577	&0.470	&0.661	&0.706	&0.685	&0.863	&0.868	&0.847	&0.858	&0.845	&\textcolor{green}{0.874}	&\textcolor{red}{0.901}	&\textbf{\textcolor{blue}{0.910}}\\
&$F_{\beta}^w\uparrow$			&0.301	&0.372	&0.173	&0.516	&0.484	&0.397	&0.694	&0.731	&0.656	&0.745	&0.718	&\textcolor{green}{0.794}	&\textcolor{red}{0.849}	&\textbf{\textcolor{blue}{0.856}}\\
&$F_{\beta}\uparrow$	&0.689	&\textcolor{red}{0.857}	&0.228	&0.588	&0.583	&0.566	&0.765	&0.800	&0.750	&0.782	&0.763	&\textcolor{green}{0.819}	&\textcolor{red}{0.857}	&\textbf{\textcolor{blue}{0.861}}\\
&$MAE\downarrow$		&0.289	&0.097	&0.194	&0.102	&0.119	&0.130	&0.055	&0.050	&0.064	&0.045	&0.049	&\textcolor{green}{0.037}	&\textcolor{red}{0.029}	&\textbf{\textcolor{blue}{0.027}}\\
									\bottomrule
	\end{tabular}}
	
	\label{tab:comparison3D}
		\vspace{-0.5cm}
\end{table*}

\subsection{Comparison with State-of-the-arts}
\label{subsection:comparison}
We compare results from our method with various state-of-the-art approaches on seven public datasets.
For fair comparisons, the results from competing methods are generated by authorized codes or directly provided by authors.

\begin{table}[t]
	\vspace{-0.1cm}
	\caption{Quantitative comparisons with state-of-the-art 2D methods.}
	\centering
	\resizebox{!}{1.87cm}{
	\begin{tabular}{lllccccccccc}
		\toprule
		 \textbf{Dataset} 	&\textbf{Metric} 			&\textbf{DSS} 		&\textbf{Amulet} 	&\textbf{R$^3$Net} 	&\textbf{PiCANet} 	&\textbf{PAGRN} 	&\textbf{EGNet} 	&\textbf{PoolNet} 	&\textbf{BASNet} 	&\textbf{CPD}  		&\textbf{Ours}\\
				&				&\cite{2DDSS}	&\cite{2DAmulet}		&\cite{2DR3Net}	&\cite{2DPiCANet}		&\cite{2DPAGRN}	&\cite{2DEGNet}		&\cite{2DPoolNet}	&\cite{2DBASNet}	&\cite{2DCPD}	&\\
			\midrule\midrule
									 
\multirow{2}{*}{\textbf{NJUD}~\cite{3DACSD}}		
&$S\uparrow$							&0.807	&0.843	&0.837	&0.847	&0.829	&0.871	&\textcolor{green}{0.872}	&\textcolor{green}{0.872}	&\textcolor{red}{0.876}	&\textbf{\textcolor{blue}{0.894}}\\	
&$F_{\beta}^w\uparrow$&0.678	&0.758	&0.736	&0.768	&0.746	&0.812	&0.816	&\textcolor{red}{0.839}	&\textcolor{green}{0.834}	&\textbf{\textcolor{blue}{0.856}}\\
&$MAE\downarrow$								&0.108	&0.085	&0.092	&0.071	&0.081	&0.057	&0.057	&\textcolor{green}{0.055}	&\textcolor{red}{0.054}	&\textbf{\textcolor{blue}{0.047}}\\
\cmidrule(r){1-12}
\multirow{2}{*}{\textbf{NLPR}~\cite{3DNLPR}}	
&$S\uparrow$							&0.816	&0.848	&0.798	&0.834	&0.844	&0.861	&0.867	&\textcolor{red}{0.890}	&\textcolor{green}{0.887}	&\textbf{\textcolor{blue}{0.907}}\\	
&$F_{\beta}^w\uparrow$							&0.614	&0.716	&0.611	&0.707	&0.707	&0.760	&0.771	&\textcolor{red}{0.834}	&\textcolor{green}{0.820}	&\textbf{\textcolor{blue}{0.850}}\\
&$MAE\downarrow$								&0.076	&0.062	&0.101	&0.053	&0.051	&\textcolor{green}{0.046}	&\textcolor{green}{0.046}	&\textcolor{red}{0.036}	&\textcolor{red}{0.036}	&\textbf{\textcolor{blue}{0.031}}\\

\cmidrule(r){1-12}
\multirow{2}{*}{\textbf{STEREO}~\cite{3DSTEREO}}
&$S\uparrow$							&0.841	&0.881	&0.855	&0.868	&0.851	&0.897	&\textcolor{green}{0.898}	&0.896	&\textcolor{red}{0.899}	&\textbf{\textcolor{blue}{0.908}}\\	
&$F_{\beta}^w\uparrow$							&0.718	&0.811	&0.752	&0.774	&0.792	&0.847	&0.849	&\textbf{\textcolor{blue}{0.873}}	&\textcolor{green}{0.865}	&\textcolor{red}{0.871}\\
&$MAE\downarrow$								&0.087	&0.062	&0.084	&0.062	&0.067	&\textcolor{green}{0.045}	&\textcolor{green}{0.045}	&\textcolor{red}{0.042}	&\textcolor{red}{0.042}	&\textbf{\textcolor{blue}{0.041}}\\

\bottomrule	
							 
	\end{tabular}}
	\vspace{0.21cm}
	
	\label{tab:comparison2D}
		\vspace{-0.35cm}
\end{table}

\begin{figure*}[t]
\centering 
\includegraphics [width=1\linewidth] {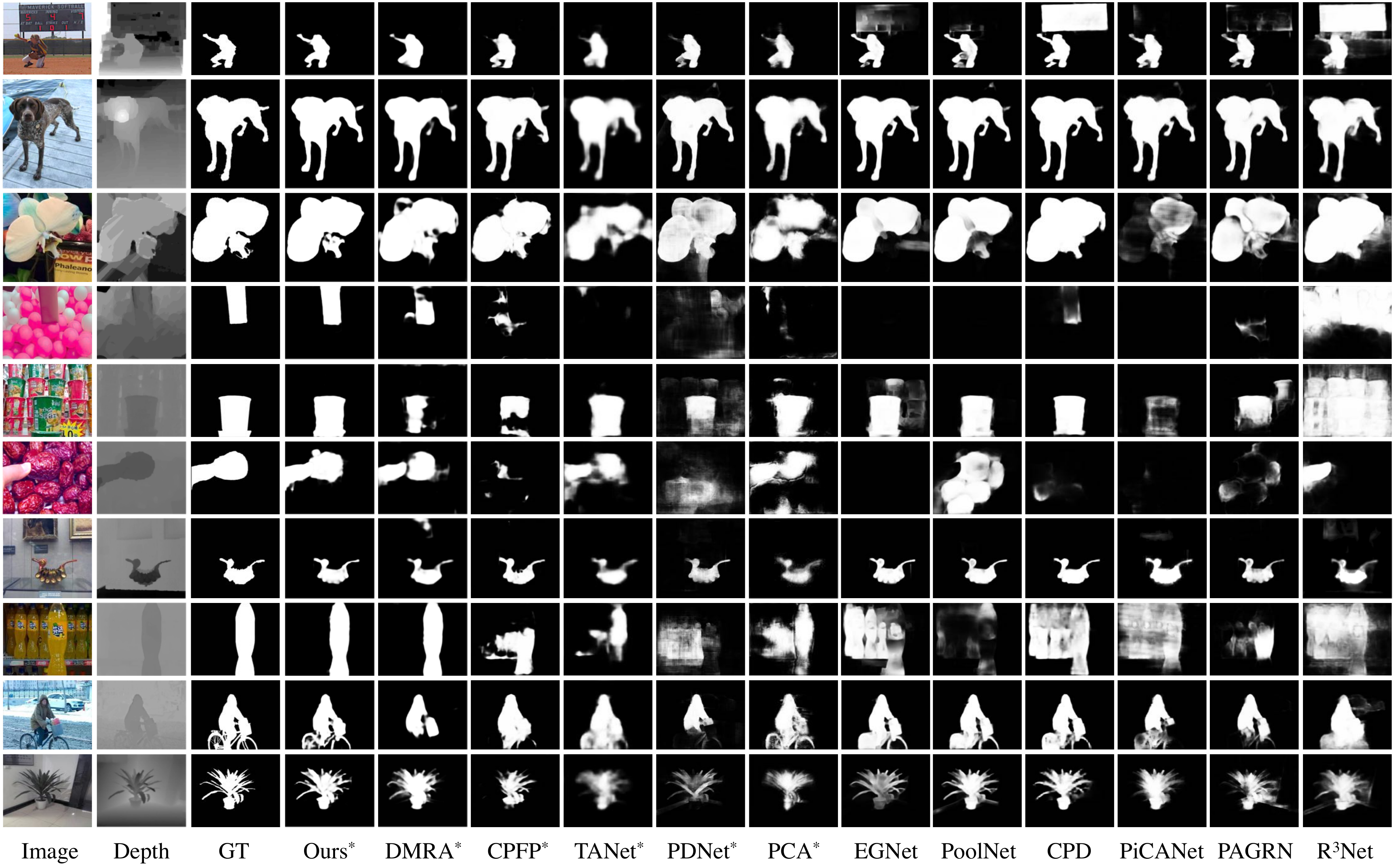}
\vspace{-0.55cm}
\caption{Visual comparisons of our method with other state-of-the-art CNNs-based methods in some representative scenes. * means RGB-D methods.}
\vspace{-0.45cm}
\label{fig:comparison}
\end{figure*}

\begin{figure*}[t]
\centering 
\includegraphics [width=1\linewidth] {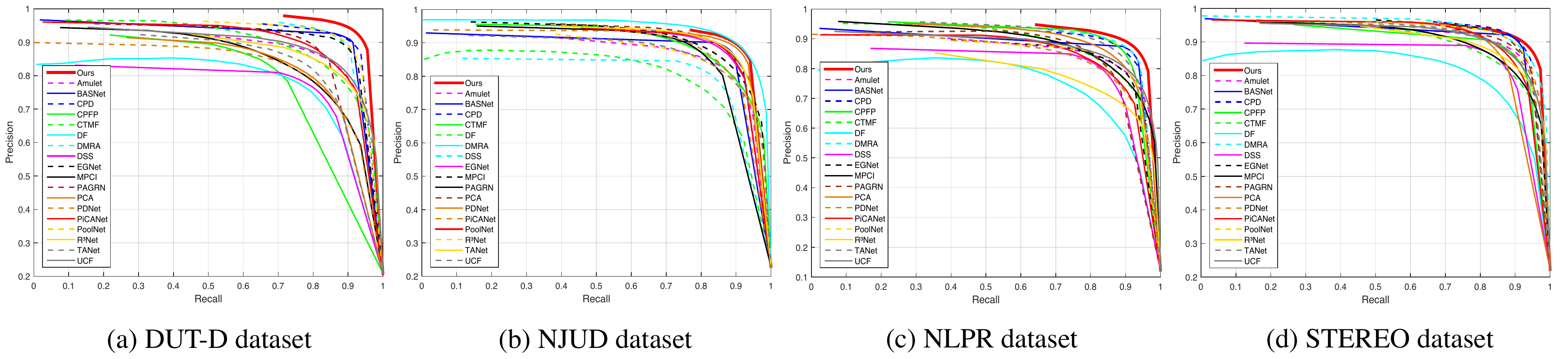}
\vspace{-0.7cm}
\caption{The PR curves of our method compared to other state-of-the-art approaches on four datasets.}
\vspace{-0.55cm}
\label{fig:PR}
\end{figure*}

\noindent{\bf Quantitative Evaluation.}
Tab.~\ref{tab:comparison3D} shows the quantitative results of our method over other 13 RGB-D ones on seven benchmark datasets.
We can see that our proposed collaborative learning framework achieves superior performance.
Noted that our method avoids the reliance on depth images and only takes RGB image as input in the testing stage.
To comprehensively verify the effectiveness of our model, we additionally conduct comparisons with 9 state-of-the-art RGB methods on three public datasets.
Results in Tab.~\ref{tab:comparison2D} consistently show that our method also achieves comparable results compared to 2D methods.
The PR curves in Fig.~\ref{fig:PR} also verify the superiority of our method.

\noindent{\bf Qualitative Evaluation.} 
Fig.~\ref{fig:comparison} shows some representative samples of results comparing our method with some top-ranking CNNs-based RGB and RGB-D approaches.
For the complex scenes with lower-contrast (the $4^{th}$ and $5^{th}$ rows) or multiple objects (the $8^{th}$ row), our method can better locate the salient objects thanks to the useful spatial information in depth image and sufficient extraction and utilization of edge information.
Thus, our method can produce accurate saliency results with sharp boundaries preserved.

\noindent{\bf Complexity Evaluation.}
We also compare the model size and run time (Frame Per Second, FPS) of our method with 11 representative models in Tab.~\ref{tab:speed}.
Thanks to the well-designed depth learning strategy, our network is free of using extra depth networks and depth inputs to make inference.
It can also be seen that our method achieves outstanding scores with a smaller model size and higher FPS (enhances FPS by 55\% compared to current best performing RGB-D model DMRA).
Those results confirm that our model is suitable for the pre-processing task in terms of model size and running speed.

\section{Conclusion}
In this work, we propose a novel collaborative learning framework for accurate RGB-D salient object detection.
In our framework, three mutually beneficial collaborators, i.e., edge detection, coarse salient object detection and depth estimation, jointly accomplish the SOD task from different perspectives.
Benefiting from the well-designed mutual-benefit learning strategy between three collaborators, our method can produce accurate saliency results with sharp boundaries preserved.
Free of using extra depth subnetworks and depth inputs during testing also makes our network more lightweight and versatile.
Experiment results on seven benchmark datasets show that our method achieves superior performance over 22 state-of-the-art RGB and RGB-D methods.

\begin{table}[t]
\vspace{0.33cm}
	\centering
	\caption{Complexity comparisons of various methods. The best three results are shown in \textbf{\textcolor{blue}{blue}}, \textcolor{red}{red}, and \textcolor{green}{green} fonts respectively. FPS means frame per second.}
	\resizebox{!}{2.3cm}{
	\begin{tabular}{ccccccccc}
		\toprule
		\multicolumn{1}{c}{} &\multicolumn{1}{c}{}  &\multicolumn{1}{c}{} &\multicolumn{1}{c}{} &\multicolumn{1}{c}{} &\multicolumn{2}{c}{NJUD~\cite{3DACSD}}   &\multicolumn{2}{c}{NLPR~\cite{3DNLPR}}\\ 
								\cmidrule(r){6-7}		 \cmidrule(r){8-9}	
		Types 				&Methods &Years &Size		&FPS  &$F_{\beta}^w\uparrow$ &$MAE\downarrow$	&$F_{\beta}^w\uparrow$ &$MAE\downarrow$\\ 
			\midrule
		\multirow{5}{*}{2D}		&DSS	&2017'CVPR	&447.3MB						&22		&0.678	&0.108	&0.614	&0.076\\
							&Amulet	&2017'ICCV	&\textbf{\textcolor{blue}{132.6 MB}}	&16		&0.758	&0.085	&0.716	&0.062\\
							&PiCANet	&2018'CVPR	&197.2 MB					&7		&0.768	&0.071	&0.707	&0.053\\
							&PoolNet	&2019'CVPR	&278.5 MB	&\textcolor{green}{32}		&0.816	&0.057	&0.771	&0.046\\
							&CPD	&2019'CVPR	&{\textcolor{green}{183 MB}}		&\textbf{\textcolor{blue}{62}}		&0.834	&0.054	&{\textcolor{green}{0.820}}	&{\textcolor{red}{0.036}}\\
							\cmidrule(r){1-9}
		\multirow{6}{*}{3D}		&PCA	&2018'CVPR	&533.6 MB	&15		&0.811	&0.059	&0.772	&0.044\\
							&TANet	&2019'TIP		&951.9 MB	&14		&0.812	&0.061	&0.789	&\textcolor{green}{0.041}\\
							&MPCI	&2019'PR		&929.7 MB	&19		&0.749	&0.079	&0.688	&0.059\\
							&PDNet	&2019'ICME	&192 MB		&19		&0.798	&0.062	&0.659	&0.064\\
							&CPFP	&2019'CVPR	&278 MB		&6		&\textcolor{green}{0.837}	&\textcolor{green}{0.053}	&\textcolor{green}{0.820}	&{\textcolor{red}{0.036}}\\
							&DMRA	&2019'ICCV	&238.8 MB	&22		&\textcolor{red}{0.853}	&{\textcolor{red}{0.051}}	&\textcolor{red}{0.845}	&\textbf{\textcolor{blue}{0.031}}	\\
							\cmidrule(r){1-9}
		\multirow{1}{*}{*}		&Ours	&			&\textcolor{red}{167.6 MB}	&\textcolor{red}{34}	&\textbf{\textcolor{blue}{0.856}}	&\textbf{\textcolor{blue}{0.047}}	&\textbf{\textcolor{blue}{0.850}}	&\textbf{\textcolor{blue}{0.031}}	\\ 								\bottomrule
	\end{tabular}}
	
	\label{tab:speed}
		\vspace{-0.55cm}
\end{table}

\section*{Acknowledgements}
This work was supported by the Science and Technology Innovation Foundation of Dalian (2019J12GX034), the National Natural Science Foundation of China (61976035), and the Fundamental  Research Funds for the Central Universities (DUT19JC58, DUT20JC42).

%
%
\bibliographystyle{splncs04}
\bibliography{egbib}

\begin{thebibliography}{10}
\providecommand{\url}[1]{\texttt{#1}}
\providecommand{\urlprefix}{URL }
\providecommand{\doi}[1]{https://doi.org/#1}

\bibitem{Fmeasure}
{Achanta}, R., {Hemami}, S.S., {Estrada}, F.J., {Süsstrunk}, S.:
  Frequency-tuned salient region detection. In: CVPR. pp. 1597--1604 (2009)

\bibitem{2DF}
{Borji}, A., {Cheng}, M.M., {Jiang}, H., {Li}, J.: Salient object detection: A
  benchmark. TIP  \textbf{24}(12),  5706--5722 (2015)

\bibitem{Benchmark}
{Borji}, A., {Sihite}, D.N., {Itti}, L.: Salient object detection: a benchmark.
  In: ECCV. pp. 414--429 (2012)

\bibitem{Canny}
{Canny}, J.: A computational approach to edge detection. TPAMI  \textbf{8}(6),
  679--698 (1986)

\bibitem{3DPCA}
{Chen}, H., {Li}, Y.: Progressively complementarity-aware fusion network for
  rgb-d salient object detection. In: CVPR. pp. 3051--3060 (2018)

\bibitem{3DTANet}
{Chen}, H., {Li}, Y.: Three-stream attention-aware network for rgb-d salient
  object detection. TIP  \textbf{28}(6),  2825--2835 (2019)

\bibitem{3DMPCI}
{Chen}, H., {Li}, Y., {Su}, D.: Multi-modal fusion network with multi-scale
  multi-path and cross-modal interactions for rgb-d salient object detection.
  PR  \textbf{86},  376--385 (2019)

\bibitem{DeepLab1}
{Chen}, L.C., {Papandreou}, G., {Kokkinos}, I., {Murphy}, K., {Yuille}, A.L.:
  Deeplab: Semantic image segmentation with deep convolutional nets, atrous
  convolution, and fully connected crfs. TPAMI  \textbf{40}(4),  834--848
  (2018)

\bibitem{DeepLab3}
{Chen}, L.C., {Zhu}, Y., {Papandreou}, G., {Schroff}, F., {Adam}, H.:
  Encoder-decoder with atrous separable convolution for semantic image
  segmentation. In: ECCV. pp. 833--851 (2018)

\bibitem{2DT2}
{Cheng}, M.M., {Zhang}, G.X., {Mitra}, N.J., {Huang}, X., {Hu}, S.M.: Global
  contrast based salient region detection. TPAMI  \textbf{37}(3),  409--416
  (2011)

\bibitem{3DDES}
{Cheng}, Y., {Fu}, H., {Wei}, X., {Xiao}, J., {Cao}, X.: Depth enhanced
  saliency detection method. In: ICIMCS. pp. 23--27 (2014)

\bibitem{3DDCMC}
{Cong}, R., {Lei}, J., {Zhang}, C., {Huang}, Q., {Cao}, X., {Hou}, C.: Saliency
  detection for stereoscopic images based on depth confidence analysis and
  multiple cues fusion. SPL  \textbf{23}(6),  819--823 (2016)

\bibitem{Robot}
{Craye}, C., {Filliat}, D., {Goudou}, J.F.: Environment exploration for
  object-based visual saliency learning. In: ICRA. pp. 2303--2309 (2016)

\bibitem{Recon2}
{Dai}, J., {Li}, Y., {He}, K., {Sun}, J.: R-fcn: object detection via
  region-based fully convolutional networks. In: NIPS. pp. 379--387 (2016)

\bibitem{2DR3Net}
{Deng}, Z., {Hu}, X., {Zhu}, L., {Xu}, X., {Qin}, J., {Han}, G., {Heng}, P.A.:
  R$^3$net: Recurrent residual refinement network for saliency detection. In:
  IJCAI. pp. 684--690 (2018)

\bibitem{2DSOC}
{Fan}, D.P., {Cheng}, M.M., {Liu}, J.J., {Gao}, S.H., {Hou}, Q., {Borji}, A.:
  Salient objects in clutter: Bringing salient object detection to the
  foreground. In: ECCV. pp. 196--212 (2018)

\bibitem{Smeasure}
{Fan}, D.P., {Cheng}, M.M., {Liu}, Y., {Li}, T., {Borji}, A.:
  Structure-measure: A new way to evaluate foreground maps. In: ICCV. pp.
  4558--4567 (2017)

\bibitem{Emeasure}
{Fan}, D.P., {Gong}, C., {Cao}, Y., {Ren}, B., {Cheng}, M.M., {Borji}, A.:
  Enhanced-alignment measure for binary foreground map evaluation. In: IJCAI.
  pp. 698--704 (2018)

\bibitem{3DSIP}
{Fan}, D.P., {Lin}, Z., {Zhao}, J., {Liu}, Y., {Zhang}, Z., {Hou}, Q., {Zhu},
  M., {Cheng}, M.M.: Rethinking rgb-d salient object detection: Models,
  datasets, and large-scale benchmarks. arXiv preprint arXiv:1907.06781  (2019)

\bibitem{videoSOD2}
{Fan}, D.P., {Wang}, W., {Cheng}, M.M., {Shen}, J.: Shifting more attention to
  video salient object detection. In: CVPR. pp. 8554--8564 (2019)

\bibitem{2DAFNet}
{Feng}, M., {Lu}, H., {Ding}, E.: Attentive feedback network for boundary-aware
  salient object detection. In: CVPR. pp. 1623--1632 (2019)

\bibitem{FastRCNN}
{Girshick}, R.: Fast r-cnn. In: ICCV. pp. 1440--1448 (2015)

\bibitem{3DCTMF}
{Han}, J., {Chen}, H., {Liu}, N., {Yan}, C., {Li}, X.: Cnns-based rgb-d
  saliency detection via cross-view transfer and multiview fusion. IEEE
  Transactions on Systems, Man, and Cybernetics  \textbf{48}(11),  3171--3183
  (2018)

\bibitem{ResNet}
{He}, K., {Zhang}, X., {Ren}, S., {Sun}, J.: Deep residual learning for image
  recognition. In: CVPR. pp. 770--778 (2016)

\bibitem{Track1}
{Hong}, S., {You}, T., {Kwak}, S., {Han}, B.: Online tracking by learning
  discriminative saliency map with convolutional neural network. In: ICML. pp.
  597--606 (2015)

\bibitem{2DDSS}
{Hou}, Q., {Cheng}, M.M., {Hu}, X., {Borji}, A., {Tu}, Z., {Torr}, P.H.S.:
  Deeply supervised salient object detection with short connections. In: CVPR.
  pp. 815--828 (2017)

\bibitem{2DT1}
{Itti}, L., {Koch}, C., {Niebur}, E.: A model of saliency-based visual
  attention for rapid scene analysis. TPAMI  \textbf{20}(11),  1254--1259
  (1998)

\bibitem{3DACSD}
{Ju}, R., {Ge}, L., {Geng}, W., {Ren}, T., {Wu}, G.: Depth saliency based on
  anisotropic center-surround difference. In: ICIP. pp. 1115--1119 (2014)

\bibitem{CRF}
{Krähenbühl}, P., {Koltun}, V.: Efficient inference in fully connected crfs
  with gaussian edge potentials. In: NIPS. pp. 109--117 (2011)

\bibitem{2DLee}
{Lee}, G., {Tai}, Y.W., {Kim}, J.: Deep saliency with encoded low level
  distance map and high level features. In: CVPR. pp. 660--668 (2016)

\bibitem{3DTPPF}
{Li}, G., {Zhu}, C.: A three-pathway psychobiological framework of salient
  object detection using stereoscopic technology. In: ICCVW. pp. 3008--3014
  (2017)

\bibitem{2DLi}
{Li}, G., {Yu}, Y.: Visual saliency based on multiscale deep features. In:
  CVPR. pp. 5455--5463 (2015)

\bibitem{2DData3}
{Li}, G., {Yu}, Y.: Visual saliency detection based on multiscale deep cnn
  features. TIP  \textbf{25}(11),  5012--5024 (2016)

\bibitem{LFS2017}
{Li}, N., {Ye}, J., {Ji}, Y., {Ling}, H., {Yu}, J.: Saliency detection on light
  field. TPAMI  \textbf{39}(8),  1605--1616 (2017)

\bibitem{2DT8}
{Li}, X., {Zhao}, L., {Wei}, L., {Yang}, M.H., {Wu}, F., {Zhuang}, Y., {Ling},
  H., {Wang}, J.: Deepsaliency: Multi-task deep neural network model for
  salient object detection. TIP  \textbf{25}(8),  3919--3930 (2016)

\bibitem{2DData1}
{Li}, Y., {Hou}, X., {Koch}, C., {Rehg}, J.M., {Yuille}, A.L.: The secrets of
  salient object segmentation. In: CVPR. pp. 280--287 (2014)

\bibitem{Segment}
{Lin}, G., {Milan}, A., {Shen}, C., {Reid}, I.D.: Refinenet: Multi-path
  refinement networks for high-resolution semantic segmentation. In: CVPR. pp.
  5168--5177 (2017)

\bibitem{2DPoolNet}
{Liu}, J.J., {Hou}, Q., {Cheng}, M.M., {Feng}, J., {Jiang}, J.: A simple
  pooling-based design for real-time salient object detection. In: CVPR. pp.
  3917--3926 (2019)

\bibitem{2DPiCANet}
{Liu}, N., {Han}, J., {Yang}, M.H.: Picanet: Learning pixel-wise contextual
  attention for saliency detection. In: CVPR. pp. 3089--3098 (2018)

\bibitem{FCN}
{Long}, J., {Shelhamer}, E., {Darrell}, T.: Fully convolutional networks for
  semantic segmentation. In: CVPR. pp. 3431--3440 (2015)

\bibitem{2DNLDF}
{Luo}, Z., {Mishra}, A.K., {Achkar}, A., {Eichel}, J.A., {Li}, S., {Jodoin},
  P.M.: Non-local deep features for salient object detection. In: CVPR. pp.
  6593--6601 (2017)

\bibitem{weightedF}
{Margolin}, R., {Zelnik-Manor}, L., {Tal}, A.: How to evaluate foreground maps.
  In: CVPR. pp. 248--255 (2014)

\bibitem{3DSTEREO}
{Niu}, Y., {Geng}, Y., {Li}, X., {Liu}, F.: Leveraging stereopsis for saliency
  analysis. In: CVPR. pp. 454--461 (2012)

\bibitem{3DNLPR}
{Peng}, H., {Li}, B., {Xiong}, W., {Hu}, W., {Ji}, R.: Rgbd salient object
  detection: A benchmark and algorithms. In: ECCV. pp. 92--109 (2014)

\bibitem{2DT6}
{Perazzi}, F., {Krähenbühl}, P., {Pritch}, Y., {Hornung}, A.: Saliency
  filters: Contrast based filtering for salient region detection. In: CVPR. pp.
  733--740 (2012)

\bibitem{3DDMRA}
{Piao}, Y., {Ji}, W., {Li}, J., {Zhang}, M., {Lu}, H.: Depth-induced
  multi-scale recurrent attention network for saliency detection. In: ICCV
  (2019)

\bibitem{2DBASNet}
{Qin}, X., {Zhang}, Z., {Huang}, C., {Gao}, C., {Dehghan}, M., {Jagersand}, M.:
  Basnet: Boundary-aware salient object detection. In: CVPR. pp. 7479--7489
  (2019)

\bibitem{3DDF}
{Qu}, L., {He}, S., {Zhang}, J., {Tian}, J., {Tang}, Y., {Yang}, Q.: Rgbd
  salient object detection via deep fusion. TIP  \textbf{26}(5),  2274--2285
  (2017)

\bibitem{3DGP}
{Ren}, J., {Gong}, X., {Yu}, L., {Zhou}, W., {Yang}, M.Y.: Exploiting global
  priors for rgb-d saliency detection. In: CVPRW. pp. 25--32 (2015)

\bibitem{Recon1}
{Ren}, S., {He}, K., {Girshick}, R.B., {Sun}, J.: Faster r-cnn: towards
  real-time object detection with region proposal networks. In: NIPS.
  vol.~2015, pp. 91--99 (2015)

\bibitem{VGG}
{Simonyan}, K., {Zisserman}, A.: Very deep convolutional networks for
  large-scale image recognition. In: ICLR (2015)

\bibitem{Track2}
{Smeulders}, A.W.M., {Chu}, D.M., {Cucchiara}, R., {Calderara}, S., {Dehghan},
  A., {Shah}, M.: Visual tracking: An experimental survey. TPAMI
  \textbf{36}(7),  1442--1468 (2014)

\bibitem{2DWang}
{Wang}, L., {Lu}, H., {Ruan}, X., {Yang}, M.H.: Deep networks for saliency
  detection via local estimation and global search. In: CVPR. pp. 3183--3192
  (2015)

\bibitem{Towards}
{Wang}, P., {Shen}, X., {Lin}, Z., {Cohen}, S., {Price}, B., {Yuille}, A.:
  Towards unified depth and semantic prediction from a single image. In: CVPR.
  pp. 2800--2809 (2015)

\bibitem{2DDeep}
{Wang}, W., {Lai}, Q., {Fu}, H., {Shen}, J., {Ling}, H.: Salient object
  detection in the deep learning era: An in-depth survey. arXiv preprint
  arXiv:1904.09146  (2019)

\bibitem{2DwangDVAP}
{Wang}, W., {Shen}, J.: Deep visual attention prediction. TIP  \textbf{27}(5),
  2368--2378 (2018)

\bibitem{2DwangFP}
{Wang}, W., {Shen}, J., {Dong}, X., {Borji}, A.: Salient object detection
  driven by fixation prediction. In: CVPR. pp. 1711--1720 (2018)

\bibitem{videoSOD1}
{Wang}, W., {Shen}, J., {Xie}, J., {Cheng}, M.M., {Ling}, H., {Borji}, A.:
  Revisiting video saliency prediction in the deep learning era. TPAMI pp.~1--1
  (2019)

\bibitem{CBAM}
{Woo}, S., {Park}, J., {Lee}, J.Y., {Kweon}, I.S.: Cbam: Convolutional block
  attention module. In: ECCV. pp. 3--19 (2018)

\bibitem{2DCPD}
{Wu}, Z., {Su}, L., {Huang}, Q.: Cascaded partial decoder for fast and accurate
  salient object detection. In: CVPR. pp. 3907--3916 (2019)

\bibitem{2DData2}
{Yan}, Q., {Xu}, L., {Shi}, J., {Jia}, J.: Hierarchical saliency detection. In:
  CVPR. pp. 1155--1162 (2013)

\bibitem{Dilation}
{Yu}, F., {Koltun}, V.: Multi-scale context aggregation by dilated
  convolutions. In: ICLR (2016)

\bibitem{2DAmulet}
{Zhang}, P., {Wang}, D., {Lu}, H., {Wang}, H., {Ruan}, X.: Amulet: Aggregating
  multi-level convolutional features for salient object detection. In: ICCV.
  pp. 202--211 (2017)

\bibitem{2DPAGRN}
{Zhang}, X., {Wang}, T., {Qi}, J., {Lu}, H., {Wang}, G.: Progressive attention
  guided recurrent network for salient object detection. In: CVPR. pp. 714--722
  (2018)

\bibitem{3DCPFP}
{Zhao}, J., {Cao}, Y., {Fan}, D., {Cheng}, M., {LI}, X., {Zhang}, L.: Contrast
  prior and fluid pyramid integration for rgbd salient object detection. In:
  CVPR (2019)

\bibitem{2DEGNet}
{Zhao}, J., {Liu}, J., {Fan}, D.P., {Cao}, Y., {Yang}, J., {Cheng}, M.M.:
  Egnet: Edge guidance network for salient object detection. In: ICCV (2019)

\bibitem{2DZhao}
{Zhao}, R., {Ouyang}, W., {Li}, H., {Wang}, X.: Saliency detection by
  multi-context deep learning. In: CVPR. pp. 1265--1274 (2015)

\bibitem{3DPDNet}
{Zhu}, C., {Cai}, X., {Huang}, K., {Li}, T.H., {Li}, G.: Pdnet: Prior-model
  guided depth-enhanced network for salient object detection. In: ICME. pp.
  199--204 (2019)

\bibitem{3DMB}
{Zhu}, C., {Li}, G., {Guo}, X., {Wang}, W., {Wang}, R.: A multilayer
  backpropagation saliency detection algorithm based on depth mining. In: CAIP.
  pp. 14--23 (2017)

\bibitem{3DCDCP}
{Zhu}, C., {Li}, G., {Wang}, W., {Wang}, R.: An innovative salient object
  detection using center-dark channel prior. In: ICCVW. pp. 1509--1515 (2017)

\end{thebibliography}

\end{document}